\documentclass{article}

\usepackage{arxiv}
\usepackage{amsmath,amssymb,amsfonts}%
\usepackage{amsthm}%
\usepackage{mathrsfs}%
\usepackage[utf8]{inputenc} 
\usepackage[T1]{fontenc}    
\usepackage{hyperref}       
\usepackage{url}            
\usepackage{booktabs}       
\usepackage{amsfonts}       
\usepackage{nicefrac}       
\usepackage{microtype}      
\usepackage[title]{appendix}
\usepackage{graphicx}
\usepackage[bf]{caption}
\usepackage{tabularray}
\usepackage{array,ragged2e}
\newcolumntype{P}[1]{>{\RaggedRight\arraybackslash}p{#1}}

\usepackage{doi}
\usepackage{etoolbox}
\usepackage[natbibapa]{apacite}
\AtBeginDocument{}
\renewcommand{\APACrefnote}[1]{}

\newtoggle{bibdoi}
\newtoggle{biburl}
\makeatletter
\newsavebox{\bib@url}
\newsavebox{\bib@doi}

\undef{\APACrefURL}
\undef{\endAPACrefURL}
\undef{\APACrefDOI}
\undef{\endAPACrefDOI}

\newenvironment{APACrefURL}
  {\global\toggletrue{biburl}\lrbox\bib@url}
  {\endlrbox}

\newenvironment{APACrefDOI}
  {\global\toggletrue{bibdoi}\lrbox\bib@doi}
  {\endlrbox}

\newcommand{\printinfo}{
  \iftoggle{bibdoi}{\usebox{\bib@doi}}{\usebox{\bib@url}}
  \togglefalse{bibdoi}
}

\AtBeginEnvironment{thebibliography}{
\pretocmd{\PrintBackRefs}{%
  \iftoggle{bibdoi}
    {\iftoggle{biburl}{\unskip\unskip}{}\usebox{\bib@doi}}
    {\iftoggle{biburl}{Retrieved from \usebox{\bib@url}}}{}
  \togglefalse{bibdoi}\togglefalse{biburl}%
}{}{}}
\makeatother

\title{MaxFloodCast: Ensemble Machine Learning Model for Predicting Peak Inundation Depth and Decoding Influencing Features}

\date{} 					

\renewcommand{\headeright}{}
\renewcommand{\undertitle}{}
\renewcommand{\shorttitle}{}

\hypersetup{
pdftitle={MaxFloodCast: Ensemble Machine Learning Model for Predicting Peak Inundation Depth and Decoding Influencing Features},
pdfsubject={},
}

\begin{document}
\maketitle

\begin{center}
{\Large
Cheng-Chun Lee\textsuperscript{a,*},
Lipai Huang\textsuperscript{a},
Federico Antolini\textsuperscript{b},
Matthew Garcia\textsuperscript{c},
Andrew Juan\textsuperscript{b},
Samuel D. Brody\textsuperscript{b},
Ali Mostafavi\textsuperscript{a,b}
\par}

\bigskip
\textsuperscript{a} Urban Resilience.AI Lab, Zachry Department of Civil and Environmental Engineering,\\ Texas A\&M University, College Station, TX\\
\textsuperscript{b} Institute for a Disaster Resilient Texas,\\ Texas A\&M University, College Station, TX\\
\textsuperscript{c} Civil and Environmental Engineering,\\ Rice University, Houston, TX\\
\vspace{6pt}
\textsuperscript{*} corresponding author, email: ccbarrylee@tamu.edu
\\
\end{center}
\bigskip
\begin{abstract}
Timely, accurate, and reliable information is essential for decision-makers, emergency managers, and infrastructure operators during flood events. This study demonstrate a proposed machine learning model, \textit{MaxFloodCast}, trained on physics-based hydrodynamic simulations in Harris County, offers efficient and interpretable flood inundation depth predictions. Achieving an average R-squared of 0.949 and a Root Mean Square Error of 0.61 ft (0.19 m) on unseen data, it proves reliable in forecasting peak flood inundation depths. Validated against Hurricane Harvey and Storm Imelda, \textit{MaxFloodCast} shows the potential in supporting near-time floodplain management and emergency operations. The model's interpretability aids decision-makers in offering critical information to inform flood mitigation strategies, to prioritize areas with critical facilities and to examine how rainfall in other watersheds influences flood exposure in one area. The \textit{MaxFloodCast} model enables accurate and interpretable inundation depth predictions while significantly reducing computational time, thereby supporting emergency response efforts and flood risk management more effectively.
\end{abstract}



\section{Introduction}\label{sec1}

As flood occurrences become more intense and frequent, decision makers, emergency managers, infrastructure owners and operators, and first responders have struggled to expeditiously assess and respond to flood losses in part due to a lack of timely, accurate, and reliable information on complex developed systems. Typically, physics-based hydrodynamic (i.e., hydrologic and hydraulic) models are used to compute flood hazards at particular areas of interest. However, these models could get computationally expensive and, worse, prohibitive, especially with increasing scale (both spatial and temporal), level of detail (resolution), and complexity (e.g., presence of hydraulic structures and multiple flood drivers). Minimizing the computational cost while maintaining the model’s accuracy, reliability, and intepretibility is therefore paramount for near-time flood inundation estimation and prediction. Earlier studies had attempted to reduce the computational burden in flood warning systems by relying on real-time precipitation as an indicator of flooding potential \citep{bedientRadarBasedFloodWarning2003, sharifUseAutomatedNowcasting2006a, javelleFlashFloodWarning2010, juanDevelopingRadarBasedFlood2017} or as a trigger to estimate inundation extent from a library of pre-delineated flood hazard maps \citep{fangEnhancedRadarBasedFlood2008, panakkalEnhancedResponseIntegration2019}. Despite these efforts, the flood hazard information provided remains limited in either scope (e.g., only considering the riverine floodplain) or scale (e.g., watershed or sub-watershed extent) due to various simplifying assumptions made in these systems. In addition, the existing models have limited interpretability to decode the extent to which different features of a region shape the flood exposure. Such interpretability in examining the importance of different features is essential for characterizing regional flood exposure and formulating flood mitigation measures. 

In recent years, machine learning (ML) has emerged as a powerful tool in flood inundation predictions \citep{mosaviFloodPredictionUsing2018a, bayatApplicationMachineLearning2022, tayfurFloodHydrographPrediction2018}, with numerous studies utilizing it to assess various aspects of floods, including flood damage, flood disruptions, and other relevant applications. Specifically, \citet{houRapidForecastingUrban2021} employed a combination of machine learning algorithms, including random forest and K-nearest neighbor, alongside a hydrodynamic-based urban flood model to predict inundation area, depth, and volumes. Similarly, \citet{mottaMixedApproachUrban2021} developed a flood prediction system that integrated machine learning classifiers and GIS techniques to identify flood-prone areas. However, a limitation of their work is the lack of determination of the contribution of different features to the prediction, which is critical for interpreting model outputs accurately. Addressing this important aspect, \citet{zahuraPredictingCombinedTidal2022} utilized Random Forest to create a surrogate model trained on environmental features from various flood events, predicting flood extent and depth in an urban coastal watershed while exploring the contribution of different physical features to localized flooding. However, a significant gap in their study is the lack of consideration for interactions from nearby and upstream areas, which is crucial in achieving accurate flood prediction. To address this gap, some researchers have explored the use of deep learning and spatial reduction and reconstruction methods \citep{leiUrbanFloodModeling2021, zhouRapidFloodInundation2021}. However, while these complex models may offer improved predictive capabilities, they often come at the cost of reduced interpretability. Recent studies (e.g., \citet{farahmandSpatialTemporalGraph2023, yuanSpatiotemporalGraphConvolutional2022}) have shown the potential of using interpretable deep learning models for evaluating feature importance in flood nowcasting. However, these studies are limited in evaluating flood status in neighborhoods as a binary state (flooded or non-flooded) and do not enable predicting peak flood depth which is essential information for evaluating the extent of risk and expected damage. In light of these challenges and advancements in the field, this paper introduces a surrogate machine learning modeling framework (called \textit{MaxFloodCast}) with novel feature engineering, providing real-time inundation predictions while taking into account nearby and upstream precipitation information, all while maintaining model interpretability. The \textit{MaxFloodCast} model is trained and tested using physics-based simulations to address concerns regarding the scarcity of historical inundation data \citep{yanRapidPredictionModel2021}. The validation of the \textit{MaxFloodCast} model in the context of two recent flood events in Harris County (Texas, USA) shows the superior performance of the model in predicting peak inundation compared with physics-based models. Thus,  the \textit{MaxFloodCast} model could perform similarly to standalone hydrodynamic models and potentially replace them in various applications, including regional floodplain management and real-time emergency operations. 

This study demonstrates the application of the \textit{MaxFloodCast} model in Harris County (Texas, USA), one of the most flood-prone areas in the United States. The validation of the model is performed in the context of the flooding caused by the 2017 Hurricane Harvey and the 2019 Storm Imelda. The \textit{MaxFloodCast} model and findings offer valuable contributions to researchers and diverse stakeholders. Machine learning models, when trained with physics-based simulations, can serve as efficient surrogate models for physics-based inundation models to improve computational speed and cost, as well as model interpretability. This capability of the \textit{MaxFloodCast} model presents two significant contributions: (1) the reduced computational cost and near-time prediction capabilities facilitate effective emergency response and provide valuable insights to emergency managers and public officials for estimating property damages, as well as identifying road inundation and disrupted access to facilities based on accurate and near-time flood inundation predictions; (2) The evaluation of features that shape inundation extent in different areas offers insights to flood managers and urban planners into flood mitigation strategies that would reduce local and regional peak inundation depths in future events. By integrating the strengths of machine learning and physics-based models, the \textit{MaxFloodCast} model could offer an efficient approach to flood peak inundation prediction and feature evaluation, enabling more effective floodplain management and informed decision-making processes for flood risk mitigation.

\section{Results}\label{sec2}
\begin{figure}[]%
\centering
\includegraphics[width=0.98\textwidth]{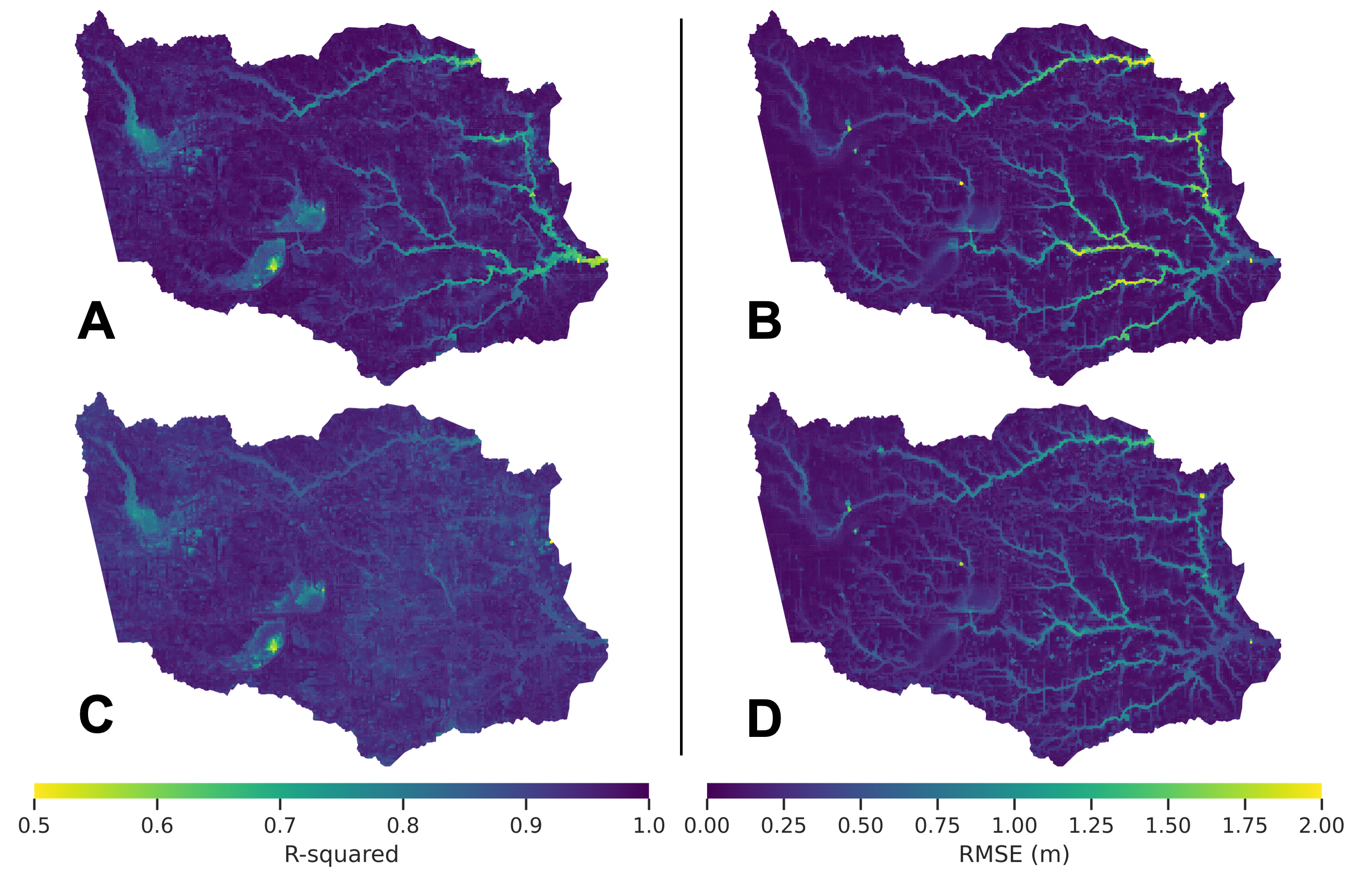}
\caption{Test R-squared and RMSE of Experiments 1 (A and B) and 2 (C and D).}\label{fig1}
\end{figure}

The study domain includes the majority of Harris County, Texas (USA), represented by 26,301 cells (details provided in Section \ref{subsec4.1}). The cells are further distinguished between “channel” and “non-channel” cells. Channel cells represent major streams/rivers (natural or manmade),  while non-channel cells represent the overland areas outside of the water bodies, which could drain into nearby channel cells or be disconnected from them entirely. Since the performance of a channel cell differs from that of a non-channel cell, we trained individual machine-learning models for each cell. The analysis includes two experiment setups utilizing tree-based XGBoost methods. The first experiment setup incorporates two precipitation features: peak and cumulative precipitation within each cell. In the second setup, additional features, such as Heavy Cumulative Precipitation Ratio in different watersheds, that incorporate information about rainfall intensity and locations are introduced. The specific feature details are discussed in Section \ref{subsec4.3}. A total of 592 simulated events with varying rainfall intensity, magnitude, and duration (details provided in Section \ref{subsec4.2}) were divided into 60\% for model training, 20\% for validation, and 20\% for testing purposes. The test R-squared (R2) and Root Mean Squared Error (RMSE) of Experiments 1 and 2 are presented in Table \ref{tab1} and Fig \ref{fig1}.
\begin{table}[]
\centering
\caption{Average test R-squared and Root Mean Squared Error values of Experiments 1 and 2}\label{tab1}
\begin{tabular*}{\textwidth}{@{\extracolsep\fill}ccccccc}
\toprule%
& \multicolumn{3}{@{}c@{}}{Experiment 1} & \multicolumn{3}{@{}c@{}}{Experiment 2} \\\cmidrule{2-4}\cmidrule{5-7}%
 & Channel & Non-Channel & Overall &  Channel & Non-Channel & Overall \\
\midrule
R\textsuperscript{2} & 0.878 & 0.953 & 0.944 & 0.916 & 0.928 & 0.926\\
RMSE & \begin{tabular}[c]{@{}c@{}}2.17 ft\\ (0.66 m)\end{tabular} & \begin{tabular}[c]{@{}c@{}}0.46 ft\\ (0.14 m)\end{tabular} & \begin{tabular}[c]{@{}c@{}}0.66 ft\\ (0.20 m)\end{tabular} & \begin{tabular}[c]{@{}c@{}}1.76 ft\\ (0.54 m)\end{tabular} & \begin{tabular}[c]{@{}c@{}}0.57 ft\\ (0.17 m)\end{tabular} & \begin{tabular}[c]{@{}c@{}}0.71 ft\\ (0.22 m)\end{tabular}\\
\hline
\end{tabular*}
\end{table}

Based on the results presented in Table \ref{tab1} and Fig \ref{fig1}, Experiment 1 demonstrates strong performance in predicting inundation depth in the non-channel cells using only peak and cumulative precipitation within each cell, with an R-squared value of 0.953 and an RMSE of 0.46 ft (0.14 m). However, the performance in the channel cells is comparatively weaker, likely due to the experiment setup of not considering precipitation in contributing nearby and upstream drainage areas. In Experiment 2, including precipitation information from contributing watersheds and drainage areas led to improved results in the channel cells, with an R-squared value increasing from 0.878 to 0.916 and an RMSE decreasing from 2.17 ft (0.66 m) to 1.76 ft (0.54 m). Nonetheless, the additional features, while beneficial for improving the performance of the ML model in channel cells, slightly worsen the performance in non-channel cells, likely due to the introduction of noise and extraneous information, leading to overfitting and increased model complexity. Considering that approximately 80\% of the cells are non-channel cells, the overall performance of Experiment 2 is inferior to that of Experiment 1.
\subsection{Differences between channel and non-channel cells}\label{subsec2.1}
\begin{figure}[]%
\centering
\includegraphics[width=0.98\textwidth]{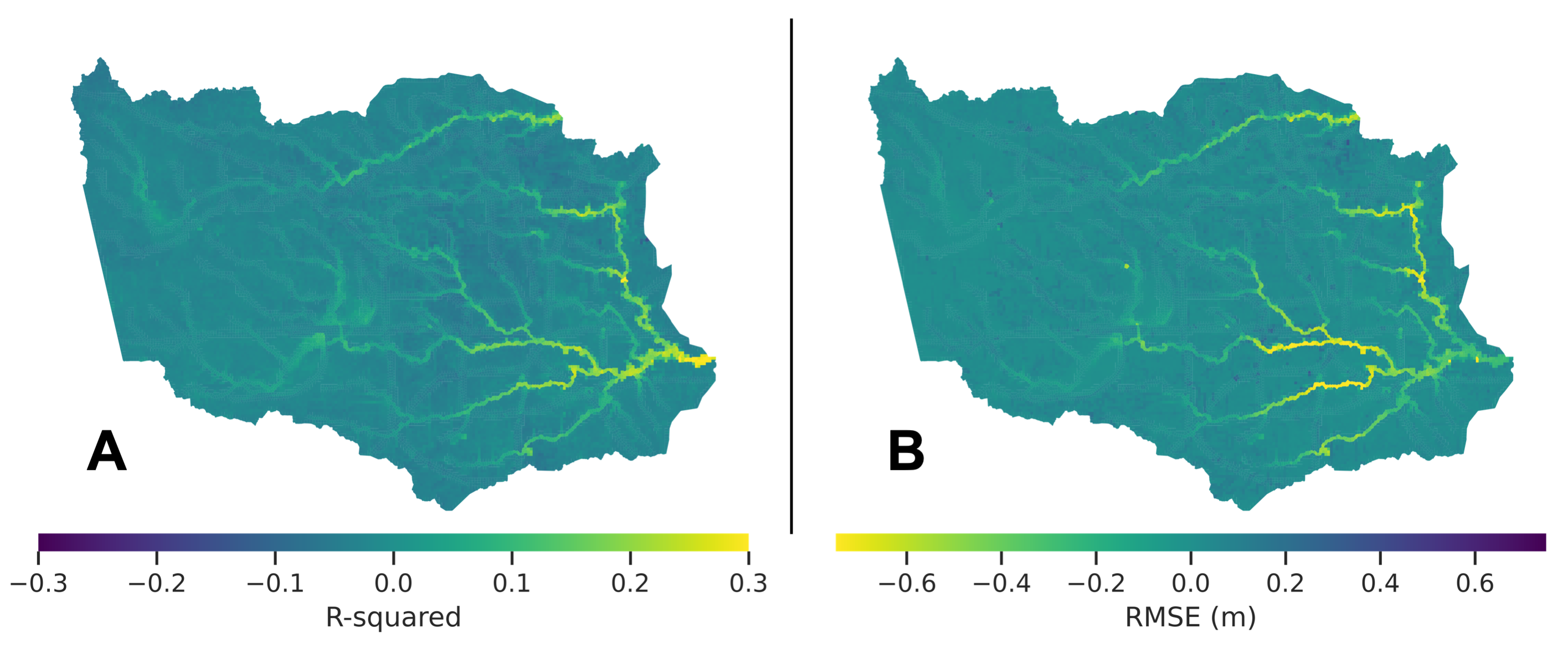}
\caption{Differences in R-squared (A) and RMSE (B) between Experiments 1 and 2, with yellow indicating improvement.}\label{fig2}
\end{figure}

Based on the results, Experiment 2 demonstrates improvements in both R-squared and RMSE values for the channel cells but worsens the performance in the non-channel cells. Fig \ref{fig2} illustrates the differences in R-squared and RMSE between Experiments 1 and 2, with yellow indicating improvement. As depicted in Fig \ref{fig2}, a significant number of channel cells have an increase of more than 0.2 in R-squared and a decrease of more than 1 ft (0.31 m) in RMSE.

These findings indicate that considering precipitation in nearby and upstream contributing drainage areas improves the prediction performance of the ML models in the channel cells. However, the inclusion of additional features results in decreased performance for the ML models in the non-channel cells. To achieve the best overall performance, Table \ref{tab2} presents the R-squared and RMSE values from Experiment 1 for the non-channel cells and Experiment 2 for the channel cells. The results indicate that the overall performance can reach an R-squared of 0.949 and an RMSE of 0.61 ft (0.19 m). These results show the capability of the \textit{MaxFloodCast} model in producing accurate and reliable peak inundation predictions in both channel and non-channel cells.
\begin{table}[]
\centering
\captionsetup{width=.95\textwidth}
\caption{Average R-squared and RMSE values of the combination of Experiments 1 and 2.}\label{tab2}
\begin{tabular}{@{}cccc@{}}
\toprule
 & Channel & Non-Channel & Overall\\
\midrule
R\textsuperscript{2}    & 0.916 & 0.953 & 0.949   \\
RMSE   &\begin{tabular}[c]{@{}c@{}}1.76 ft\\ (0.54 m)\end{tabular} & \begin{tabular}[c]{@{}c@{}}0.46 ft\\ (0.14 m)\end{tabular} & \begin{tabular}[c]{@{}c@{}}0.61 ft\\ (0.19 m)\end{tabular} \\
\hline
\end{tabular}
\end{table}
\subsection{Feature importance in the prediction of peak inundation depth}\label{subsec2.2}
\begin{figure}[]%
\centering
\includegraphics[width=0.98\textwidth]{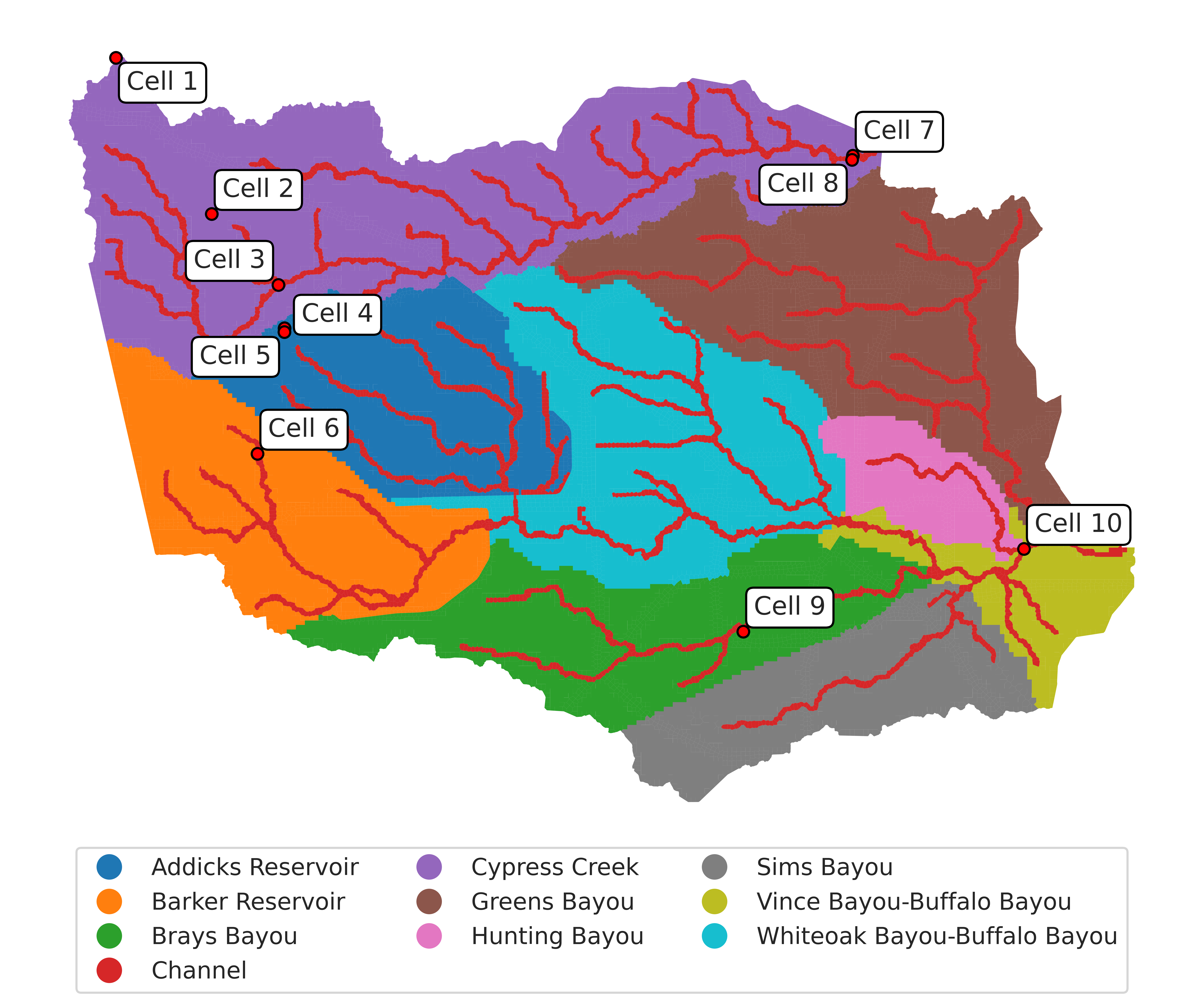}
\caption{Locations of the 10 selected cells chosen for further discussion.}\label{fig3}
\end{figure}

\begin{figure}[h]%
\centering
\includegraphics[width=0.98\textwidth]{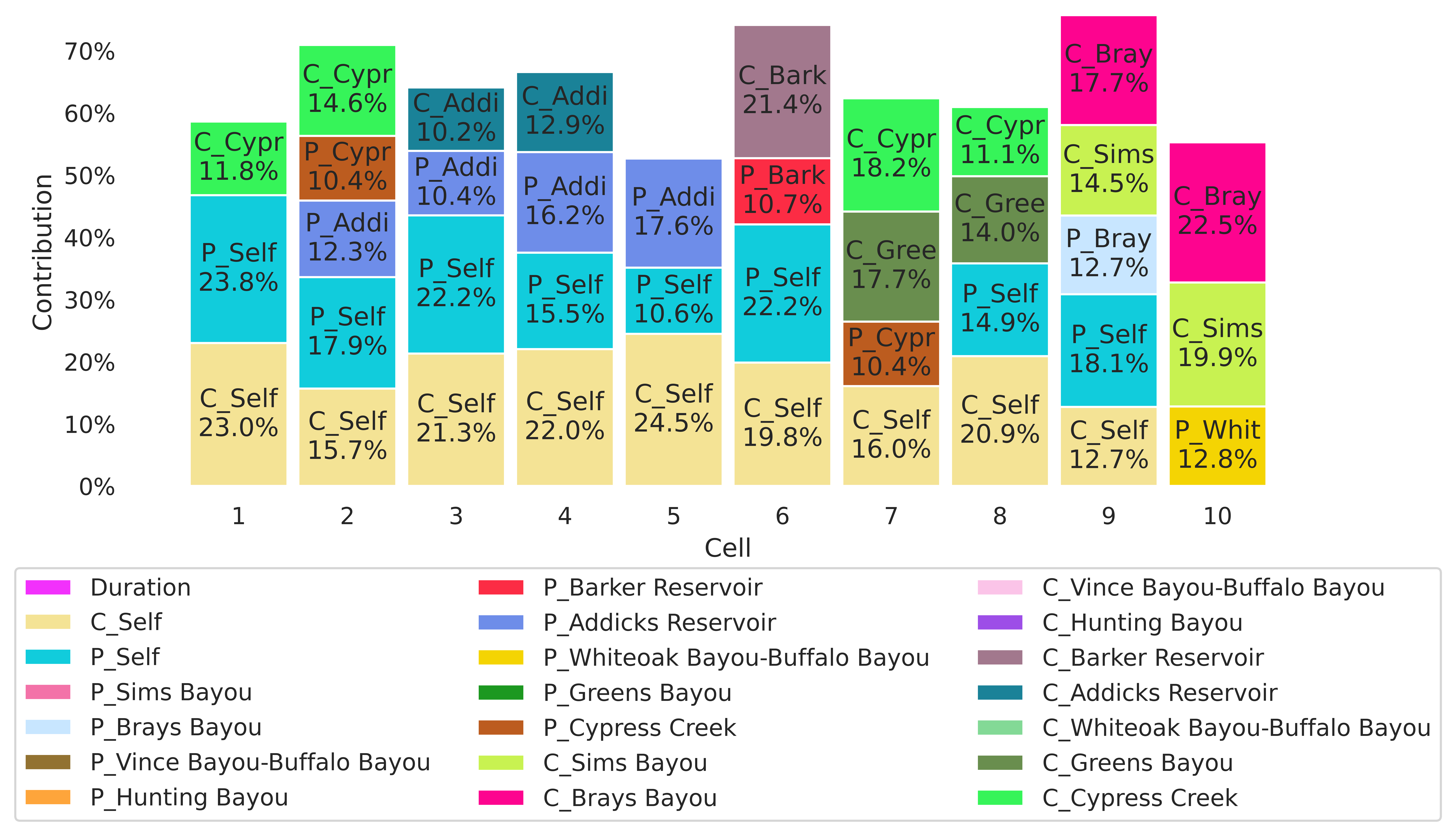}
\caption{Features that contribute more than 10\% to the depth prediction for the 10 selected cells (Note: ‘C’ and ‘P’ indicate cumulative and peak rainfall, respectively, while ‘Self’ denotes individual cells)}\label{fig4}
\end{figure}

Given that Experiment 2 incorporates 21 features (details in Section \ref{subsec4.2}), representing the precipitation status in various areas within Harris County for model training, it becomes crucial to assess whether the trained models weighed features that align with engineering judgments. We select 10 cells in the study area that represent different characteristics, including upstream/downstream and channel/non-channel cells, for further analysis. The locations and descriptions of these 10 selected cells are provided in Fig \ref{fig3} and Table \ref{tab3}, respectively.
\begin{table}[h]
\captionsetup{width=.95\textwidth}
\caption{Details of the 10 selected cells chosen for further discussion.}\label{tab3}
\centering
\begin{tabular}{@{}cP{.6\textwidth}P{.3\textwidth}@{}}
\hline
Cell & Location description & Watershed\\
\hline
1 & Study area perimeter & Cypress Creek\\
2 & Area disconnected from the channel network & Cypress Creek\\
3 & Cell containing two distinct channels differing by size, drainage, and streambed elevation & Cypress Creek\\
4 & Area on the edge of a steep terrain depression & Addicks Reservoir\\
5 & Terrain depression (quarry) & Addicks Reservoir\\
6 & Upstream channel & Barker Reservoir~\\
7 & Downstream channel & Cypress Creek\\
8 & Downstream floodplain & Cypress Creek\\
9 & Residential area in the floodplain & Brays Bayou\\
10 & Downstream channel, after a major confluence & Buffalo and Hunting Bayous\\
\hline
\end{tabular}
\end{table}
The \textit{MaxFloodCast} model is based on XGBoost method. One of the key benefits of using XGBoost is its ability to identify and prioritize essential features, consequently enhancing both model accuracy and interpretability. XGBoost provides valuable insights into each feature's contributions, enabling a better understanding of their importance in making predictions. Fig \ref{fig4} demonstrates the features that contribute more than 10\% to the prediction of inundation depth for the 10 selected cells. 
The cumulative and peak precipitations within the cells play a primary role in predicting inundation depth for most of the selected cells, except for Cells 7 and 10, both being downstream channel cells. Water depth prediction in Cell 7 is affected by the heavy cumulative and peak precipitation ratios in its watershed (Cypress Creek) and by the heavy cumulative precipitation ratio in a nearby watershed (Greens Bayou). More notably, Cell 10, which has the largest drainage of the selected cells, exhibits a contribution of both peak and cumulative precipitation within the cell that is less than 10\%. Instead, the prediction of inundation depth for Cell 10 is predominantly influenced by its upstream watersheds, specifically the heavy cumulative precipitation ratio in the Brays Bayou and Sims Bayou watersheds, as well as the heavy peak precipitation ratio in the Whiteoak Bayou-Buffalo Bayou watershed.

Comparing the downstream cells, such as Cells 7 and 10, to the upstream cells, the latter tend to rely more on the peak and cumulative precipitation information within their own cells for predicting inundation depth. For example, Cells 1 to 6 show a relatively higher contribution of peak and cumulative precipitation than Cells 7 to 10. Moreover, channel cells exhibit a greater need for information from nearby and upstream areas, while non-channel cells tend to rely more on the peak and cumulative precipitation within their own cell. For instance, Cell 8, which is close to Cell 7, relies more on precipitation information within its cell for prediction purposes. 

The findings of this study demonstrate that the trained ML models can effectively predict inundation depth by considering features that align with the physical characteristics of the study area as well as engineering judgment. The downstream cells show a tendency to rely on upstream information, whereas predicting inundation in channel cells exhibits a greater need for information from both nearby and upstream areas compared to non-channel cells. These results not only enhance the performance of ML models but also instill trust in users, such as emergency responders and managers, who can utilize the prediction results with confidence. Additionally, these results reveal interdependencies among various watersheds and the extent to which peak and cumulative rainfall in other watersheds would influence peak inundation depth in a cell in a distinct watershed. Such interpretability and evaluation of interdependencies among watersheds are unique aspects of the \textit{MaxFloodCast} model compared with the existing models.
\subsection{Model validation with Hurricane Harvey and Tropical Storm Imelda}
To validate the effectiveness of the developed ML model in predicting flood inundation depth in Harris County, we utilized two severe storm events, Hurricane Harvey and Tropical Storm Imelda, as case studies in the study area. Hurricane Harvey, occurring in August 2017, stands as one of the most devastating storms that caused catastrophic flooding in Texas, impacting the Texas Gulf Coast and resulting in extensive damage to properties and infrastructure. On the other hand, Tropical Storm Imelda struck the Houston metropolitan area in September 2019. While not as intense as Harvey, Imelda still caused widespread flooding in several regions, affecting homes, businesses, and transportation systems. These storms were selected for their contrasting characteristics, including size, rainfall distributions, and total rainfalls, allowing the ML models to be validated on both larger and more generally uniform rainfall events, such as Harvey, as well as more localized events, like Imelda. This paper utilizes gage records (see Section \ref{subsec4.2} for more details) collected during Hurricane Harvey and Tropical Storm Imelda to validate the prediction performance of the surrogate ML models. The performance of the physics-based model, HEC-RAS, that was used in the training of the surrogate ML models is also listed for reference. Tables \ref{tab4} and \ref{tab5} present the performance evaluation of both the HEC-RAS model and the two ML models, utilizing Root Mean Squared Error (RMSE) and Mean Absolute Percentage Error (MAPE) as performance metrics.
\begin{table}[]
\centering
\caption{Model validation on inundation depths versus gage records during Hurricane Harvey}\label{tab4}
\begin{tabular*}{\textwidth}{@{\extracolsep\fill}ccccccc}
\toprule%
& \multicolumn{2}{@{}P{0.2\textwidth}@{}}{Gage depth $<$ 15ft(4.57m): (n=24)} & \multicolumn{2}{@{}P{0.2\textwidth}@{}}{Gage depth $>=$ 15ft(4.57m) and $<$ 25ft(7.62m): (n=40)} & \multicolumn{2}{@{}P{0.2\textwidth}@{}}{Gage depth $>=$ 25ft(7.62m): (n=34)} \\\cmidrule{2-3}\cmidrule{4-5}\cmidrule{6-7}%
 & RMSE & MAPE & RMSE & MAPE & RMSE & MAPE \\
\midrule
HEC-RAS & \begin{tabular}[c]{@{}c@{}}3.24 ft\\ (0.99 m)\end{tabular} & 14.66\% & \begin{tabular}[c]{@{}c@{}}2.38 ft\\ (0.73 m)\end{tabular} & 8.48\% & \begin{tabular}[c]{@{}c@{}}3.48 ft\\ (1.06 m)\end{tabular} & 8.93\%\\
ML-Exp1 & \begin{tabular}[c]{@{}c@{}}2.21 ft\\ (0.67 m)\end{tabular} & 13.28\% & \begin{tabular}[c]{@{}c@{}}3.66 ft\\ (1.12 m)\end{tabular} & 13.96\% & \begin{tabular}[c]{@{}c@{}}7.25 ft\\ (2.21 m)\end{tabular} & 17.68\%\\
ML-Exp2 & \begin{tabular}[c]{@{}c@{}}2.50 ft\\ (0.76 m)\end{tabular} & 15.22\% & \begin{tabular}[c]{@{}c@{}}4.06 ft\\ (1.24 m)\end{tabular} & 16.20\% & \begin{tabular}[c]{@{}c@{}}7.62 ft\\ (2.32 m)\end{tabular} & 18.50\%\\
\hline
\end{tabular*}
\end{table}
\begin{table}[]
\centering
\caption{Model validation on inundation depths versus gage records during Tropical Storm Imelda}\label{tab5}
\begin{tabular*}{\textwidth}{@{\extracolsep\fill}ccccccc}
\toprule%
& \multicolumn{2}{@{}P{0.2\textwidth}@{}}{Gage depth $<$ 8ft(2.44m): (n=23)} & \multicolumn{2}{@{}P{0.2\textwidth}@{}}{Gage depth $>=$ 8ft(2.44m) and $<$ 15ft(4.57m): (n=40)} & \multicolumn{2}{@{}P{0.2\textwidth}@{}}{Gage depth $>=$ 15ft(4.57m): (n=37)} \\\cmidrule{2-3}\cmidrule{4-5}\cmidrule{6-7}%
 & RMSE & MAPE & RMSE & MAPE & RMSE & MAPE \\
\midrule
HEC-RAS & \begin{tabular}[c]{@{}c@{}}4.83 ft\\ (1.47 m)\end{tabular} & 236.16\% & \begin{tabular}[c]{@{}c@{}}3.59 ft\\ (1.09 m)\end{tabular} & 23.37\% & \begin{tabular}[c]{@{}c@{}}4.46 ft\\ (1.36 m)\end{tabular} & 15.72\%\\
ML-Exp1 & \begin{tabular}[c]{@{}c@{}}4.15 ft\\ (1.27 m)\end{tabular} & 126.11\% & \begin{tabular}[c]{@{}c@{}}3.68 ft\\ (1.12 m)\end{tabular} & 27.3\% & \begin{tabular}[c]{@{}c@{}}5.34 ft\\ (1.63 m)\end{tabular} & 20.84\%\\
ML-Exp2 & \begin{tabular}[c]{@{}c@{}}3.77 ft\\ (1.15 m)\end{tabular} & 158.23\% & \begin{tabular}[c]{@{}c@{}}3.26 ft\\ (0.99 m)\end{tabular} & 24.09\% & \begin{tabular}[c]{@{}c@{}}5.35 ft\\ (1.63 m)\end{tabular} & 20.01\%\\
\hline
\end{tabular*}
\end{table}
While Hurricane Harvey had larger and more uniform rainfall, with almost the entire Harris County experiencing precipitation heavier than two inches (the threshold for the features used in this paper), the additional precipitation features introduced in Experiment 2 do not lead to an improvement in prediction performance. In fact, these additional features seem to weaken the model by potentially introducing noise and extra information. Specifically, when the gage depth is less than 15 ft (4.57 m), the ML models perform similarly to the physics-based model in terms of MAPE and even outperform it in terms of RMSE. However, as the gage depth exceeds 15 ft (4.57 m), the performance of the ML models declines. This could be attributed to a limited number of physics-based simulation events at this severity level in the training data, resulting in slightly worse prediction performance.

On the other hand, during Tropical Storm Imelda, which had non-uniform, less intense rainfall, and shorter duration compared to Hurricane Harvey, the additional precipitation features in Experiment 2 slightly improve the prediction performance compared to Experiment 1. This result highlights the capability of the \textit{MaxFloodCast} model in Experiment 2 to incorporate relevant precipitation information from different watersheds when the distribution of precipitation is not uniform. Furthermore, the results demonstrate that the prediction performance of the ML models during Tropical Storm Imelda is comparable to the physics-based model when the gage depth is less than 15 ft (4.57 m) and only slightly worse when the depth exceeds 15 ft (4.57 m). This finding indicates that the \textit{MaxFloodCast} model can effectively serve as a surrogate model to the physics-based model, providing near-time predictions to support situation awareness and decision-making processes.
\section{Discussion}\label{sec4}
The study demonstrates that ML models, trained using physics-based model simulations, can provide comparable prediction performance to hydrodynamic models while improving computational efficiency and model interpretability. The \textit{MaxFloodCast} model presented in this study provides a prediction performance closely aligned with  the performance of standalone hydrodynamic models in predicting peak flood depths. The results of testing the model in Harris County (Texas, USA) highlight the promising potential of a surrogate ML modeling framework in supporting near-time floodplain management and emergency operations. The significant advantage of the \textit{MaxFloodCast} model lies in its remarkably lower computational cost compared to conventional hydrodynamic models for city-scale predictions. The \textit{MaxFloodCast} model predictions related to Hurricane Harvey and Storm Imelda were achieved in just a few seconds once the models were trained, in contrast to the flood depth calculation in a hydrodynamic model, which took more than 100 times longer. Generally, this runtime gap tends to increase with the extent of modeled area (from water-shed level to multi-watershed city-scale predictions), spatial and temporal resolution, and number of predicted features. The model can take rainfall data as input and produce peak inundation predictions for future events. Having computationally faster and reliable predictions of maximum flood inundation depth is crucial for flood managers, emergency managers, urban planners, and decision-makers planning and responding to flood events, allowing for the consideration of multiple scenarios and the implementation of mitigation measures with confidence. The model also provides valuable support for emergency response by enabling quicker estimation impacts (based on peak inundation prediction) for the allocation and staging of resources as flood events unfold. For example, the peak inundation predictions could be used to estimate which critical roads and facilities (e.g., hospitals and shelters) could be flooding due to an ongoing storm and hence plan proactively to respond to the impacts. By compressing the planning-response cycle, losses from flooding in vulnerable areas could be more effectively reduced. The model is adaptable to other regions in the U.S. and across the world, given proper training on location-specific physics-based models.

The interpretability and evaluation of watershed interdependencies are unique aspects that set the \textit{MaxFloodCast} model apart from existing flood prediction models. The feature importance analysis offers critical information to inform flood mitigation strategies, allowing decision-makers to prioritize areas with critical facilities, such as shelters and hospitals, and examine how rainfall in other watersheds influences flood exposure in one area. These insights could enable a more system-of-systems approach to flood risk assessment by examining the spillover effects of certain features in other watersheds on flood inundation status in a certain watershed. Furthermore, the model's ability to be conveniently re-trained over time based on the updated data of hydrodynamic modeling results that consider city development and flood mitigation measures enhances its practicality for future events. The adaptability of the model ensures that the model remains relevant and effective in supporting emergency management, flood mitigation efforts, and city planning.

In sum, the integration of physics-based simulations and ensemble machine learning techniques, particularly XGBoost, presents a promising and efficient framework for near-time flood prediction and management. The proposed \textit{MaxFloodCast} model enables accurate and interpretable inundation depth predictions while significantly reducing computational time, thereby supporting emergency response efforts and flood risk management more effectively.
\section{Materials and Methods}
\subsection{Study area}\label{subsec4.1}
Harris County, in Southeast Texas, is the largest county of the Greater Houston Metropolitan Statistical Area, with an area of 1,778 sq mi (4,605 sq km) and a population that grew past 4.5 million people in the past ten years. The topography is substantially flat, with elevation ranging from -40 ft (-12.19 m) up to 300 ft (91.44 m) above mean sea level in the northwest corner. More than two-thirds of the county area is classified as developed, while about 20\% is classified as pasture and cultivated in the western part of the county (\cite{dewitzNationalLandCover2021}). The two main watersheds are Cypress Creek in the north and the Buffalo Bayou system in the central and southern portions of the county. Both flow eastward into the low course of the San Jacinto River and the Ship Channel, which eventually are connected to the Gulf of Mexico. The presence of a densely built environment, slow natural drainage, and scarce soil infiltration capacity, together with its location in a subtropical climatic region, make Harris County chronically prone to flooding. Figure 5 shows the details of the study area along with the boundary of Harris County and the distribution of gages.
\begin{figure}[]%
\centering
\includegraphics[width=0.75\textwidth]{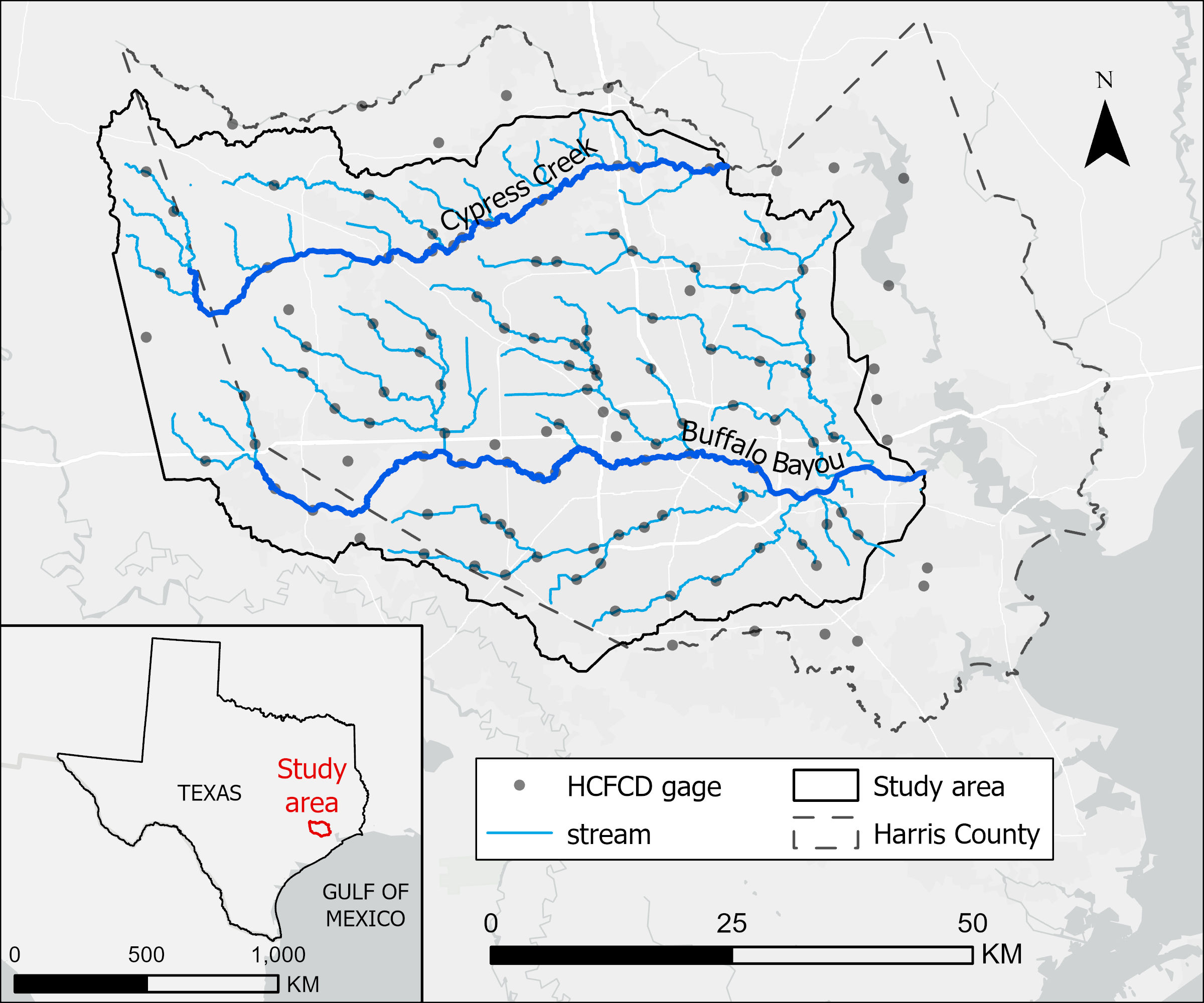}
\caption{Study area and Harris County Flood Control District gage distribution.}\label{fig5}
\end{figure}

\subsection{Ground truth and validation data (HEC-RAS and Harvey/Imeda)}\label{subsec4.2}
The ML flood inundation prediction model presented in this study is trained using physics-based simulations to address the challenges arising from the scarcity of historical inundation data. Specifically, the simulation flood inundation depths are provided by the HEC-RAS model, which is a hydrologic software developed by the Army Corps of Engineers. In this study, we built a two-dimensional model for unsteady flow simulation. The study area was covered with a 1,200 x 1,200 sq ft (approx. 366 m) mesh grid, which was further refined along the major watercourses and tributaries with computations along these centerlines to form pseudo-cross sections for stability and computational speed. Breaklines were applied in correspondence to interstates, highways, and other known high points to improve accuracy. The resulting mesh comprises 26,301 cells, providing a detailed representation of the study area. To establish the necessary parameters for the simulation, we implemented Manning's roughness and imperviousness values based on the 2019 National Land Cover Dataset \citep{dewitzNationalLandCover2021}. Additionally, soil infiltration parameters were acquired from the Gridded Soil Survey Geographic (gSSURGO) database \citep{usdanrcsGriddedSoilSurvey2023}.

To validate the HEC-RAS model, all Harris County Flood Control District gages \citep{harriscountyfloodcontroldistrictHarrisCountyFlood2023} were analyzed for completeness of data during five events spanning from 2016 to 2020. These events spanned a wide range of types, from extremely localized, like the 2016 Tax Day event, to widely uniform, like the 2017 Hurricane Harvey event. These concluded with Tropical Storm Beta in 2020, as this is when the original model was developed. The gages and events were then calibrated for global minimum error on all gages since there were significant deviations on flow runoff volumes from different parts of the domain with similar land use and soil type from antecedent moisture conditions and other similar factors that were all globally set for each event. This model was developed for a methodology paper showing that it was worth pursuing ML for these types of large computationally heavy Partial Differential Equations solvers for surrogate modeling \citep{garciaLeveragingMeshModularization2023}.

Besides HEC-RAS data, this study looks at 2017 Hurricane Harvey and 2019 Tropical Storm Imelda to validate the ML model. We leveraged Harris County Flood Control District’s network of rain and stream gages \citep{harriscountyfloodcontroldistrictHarrisCountyFlood2023}, which makes Houston one of the most gaged cities in the country \citep{sikderCaseStudyRapid2019}. We used rain gage records from 122 stations to model the rainfall. These gages have a high temporal resolution, with records every 15 minutes, and are well-distributed within and outside the study area (Fig \ref{fig5}), making them more reliable than other data sources for the use case. We modeled 7-day periods, 8/25-9/1/2017 for Hurricane Harvey and 9/15-9/22/2019 for Tropical Storm Imelda, in HEC-RAS using the Thiessen polygons method. We then extracted peak hourly intensity, cumulative precipitation, and event duration values at the cell level, which feed the ML model. Later, we compared ML model results against peak water surface elevation recorded or reconstructed at the same stations \citep{harriscountyfloodcontroldistrictHarrisCountyFlood2023}.
\subsection{Precipitation data and corresponding feature engineering} \label{subsec4.3}
This paper employs rainfall data as input for machine learning models, serving as a surrogate model for physics-based hydrodynamic models. Rainfall data provides 3d precipitation information defined by latitude, longitude, and time. With 26,301 cells in the study area, we utilize interpolation to estimate the hourly rainfall intensity within each cell. This approach treats the precipitation of each cell as a time series, enabling the extraction of features such as cumulative and peak precipitations by aggregating and identifying the maximum values within the time series.

In addition to utilizing the cumulative and peak precipitation data within each cell, heavy peak/cumulative precipitation is mapped based on a threshold of two inches (50.8 mm). Binary codes are generated for each cell, with a value of 1 indicating heavy peak/cumulative precipitation areas and 0 for other areas. This intermediary feature enhances the understanding of precipitation patterns in different areas, thereby improving the overall predictive modeling performance. Specifically, the paper engineers two essential features: Heavy Cumulative Precipitation Ratio and Heavy Peak Precipitation Ratio.

\begin{figure}[]%
\centering
\includegraphics[width=0.98\textwidth]{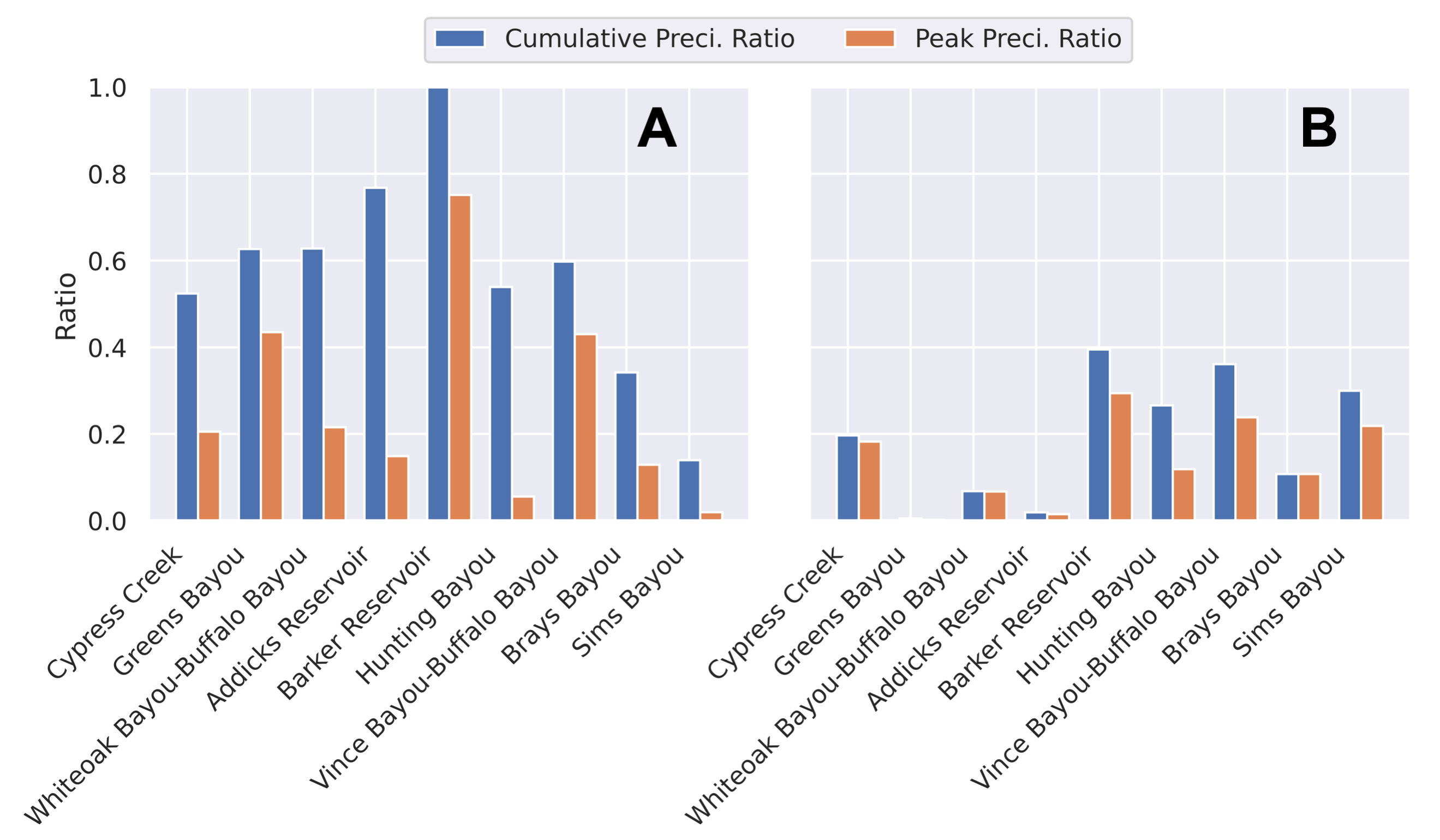}
\caption{Heavy cumulative and peak precipitation ratios of two example events in different watersheds.}\label{fig6}
\end{figure}

To achieve this, the study divides Harris County into nine distinct watershed regions, as depicted in Fig \ref{fig3}. For each watershed region, the Heavy Cumulative Precipitation Ratio and Heavy Peak Precipitation Ratio are calculated based on the identified heavy precipitation cells. The calculations for both ratios are identical, except for the distinction in the identification of heavy precipitation based on different precipitation categories (cumulative and peak). The following equations represent the calculation of the Heavy Cumulative Precipitation Ratio and Heavy Peak Precipitation Ratio for a given Watershed Region $i$ in this study:
\begin{equation}
\text{Heavy Peak/Cumulative Precipitation Ratio}_i = \frac{\sum_{j}^{}AC_{i,j}\times h_{i,j}}{AW_{i}}
\end{equation}
\begin{equation}
    h_{i, j} =
        \begin{cases}
            1, & \text{if } p_{i, j} > 2 \text{ inches}, \\
            0, & \text{otherwise}.
        \end{cases}
\end{equation}
where $AW_i$ represents the area of Watershed Region $i$, $AC_{i, j}$ represents the area of Cell $j$ in Watershed Region $i$, $h_{i, j}$ is a binary identifier of Cell $j$ in Watershed Region $i$, and $p_{i, j}$ denotes the peak/cumulative precipitation in Cell $j$ within Watershed Region $i$. The calculated Heavy Cumulative Precipitation Ratio and Heavy Peak Precipitation Ratio remain constant across all cells for a given event. These two ratios can vary significantly across different events and watershed regions, providing substantial explanatory potential. For instance, as illustrated in Fig \ref{fig6}A, a higher Heavy Cumulative Precipitation Ratio is not necessarily associated with a higher Heavy Peak Precipitation Ratio. On the other hand, Fig \ref{fig6}B demonstrates that while the Heavy Cumulative Precipitation Ratio and Heavy Peak Precipitation Ratio are correlated, the precipitation statuses in different watershed regions are distinct. These observations highlight the importance of considering both ratios to gain comprehensive insights into precipitation characteristics during various events and across diverse geographic areas. Table \ref{tab6} shows all features utilized in this paper for training the ML model to predict flood inundation depth.
\begin{table}[h]
\centering
\caption{Features utilized in this study for training ML models}
\label{tab6}
\begin{tabular}{@{}P{0.3\textwidth}P{0.2\textwidth}P{0.35\textwidth}@{}}
\hline
Feature & Unit & Description \\ \hline
Cumulative precipitation & Inch & Summation of event precipitation \\
Peak precipitation & Inch & Max event precipitation \\
Duration & Hour & Precipitation duration \\
Heavy cumulative precipitation ratios & Scalar (ranged 0 to 1) & Ratio of heavy cumulative precipitation in a watershed area to the entire watershed area. \\
Heavy peak precipitation ratios & Scalar (ranged 0 to 1) & Ratio of heavy peak precipitation in a watershed area to the entire watershed area. \\ \hline
\end{tabular}
\end{table}
\subsection{Machine learning method: XGBoost}\label{subsec4.4}
XGBoost has demonstrated excellence in various fields, including civil infrastructure management \citep{amjadPredictionPileBearing2022, dongXGBoostAlgorithmbasedPrediction2020, macaulayMachineLearningTechniques2022} and flood risk management \citep{yuanPredictingRoadFlooding2021, maXGBoostbasedMethodFlash2021}. Its ability to handle large datasets and capture intricate relationships enables accurate predictions and informed decision-making \citep{maXGBoostbasedMethodFlash2021}. The robustness and efficiency of XGBoost make it suitable for real-time applications, providing valuable insights for disaster preparedness and response.

In this study, 592 simulation events were partitioned into training (356 events), validation (118 events), and test (118 events) datasets. In addition, all features were rescaled between 0 and 1 using a constant feature scale derived from the training data. The objective function is to minimize squared errors, with a learning rate of 0.01. The model constructed 1000 trees with a maximum depth of 5 to balance complexity and performance. L1 regularization was applied to prevent overfitting, and 'colsample\_bytree' was set to 0.3 for additional randomness.

The study involved two experiment setups. The first incorporated peak and cumulative precipitation within each cell. The second setup added Heavy Cumulative Precipitation Ratio and Heavy Peak Precipitation Ratios in different watersheds, providing information on rainfall intensity and locations. The XGBoost models were trained on 20 CPUs for parallel computation, with each experiment taking approximately three hours. The model checkpoint was stored at the optimal validation RMSE for each cell and used for flood depth prediction in the case study.

\section{Data Availability}
The data that support the findings of this study is available from the corresponding author upon request.

\section{Code Availability}
The code that supports the findings of this study is available from the corresponding author upon request.

\section{Acknowledgements}
The authors would like to acknowledge funding support from the National Science Foundation under CRISP 2.0 Type 2, Grant 1832662, and the Texas A\&M X-Grant Presidential Excellence Fund. Any opinions, findings, conclusions, or recommendations expressed in this research are those of the authors and do not necessarily reflect the view of the funding agencies.

\bibliography{ref}

\end{document}